\documentclass[conference]{IEEEtran}
\IEEEoverridecommandlockouts
\usepackage{cite}
\usepackage{amsmath,amssymb,amsfonts}
\usepackage{algorithmic}
\usepackage{graphicx}
\usepackage{textcomp}
\usepackage{xcolor}

\usepackage{multirow} 
\usepackage{float} 
\usepackage{array} 

\def\BibTeX{{\rm B\kern-.05em{\sc i\kern-.025em b}\kern-.08em
    T\kern-.1667em\lower.7ex\hbox{E}\kern-.125emX}}
\begin{document}

\title{Project URSULA: Design of a Robotic Squid for Underwater Manipulation\\

\thanks{This research is supported by The Scientific and Technological Research Council of Turkey (T\"{U}B\.{I}TAK) through the project entitled ``Robotic Squid for Underwater Manipulation and Intervention'' (grant no. 216M201).}
}

\author{\IEEEauthorblockN{Berke Gur}
\IEEEauthorblockA{\textit{Department of Mechanical Engineering} \\
\textit{Marmara University}\\
\.{I}stanbul, T\"{u}rkiye \\
berke.gur@marmara.edu.tr}
}

\maketitle

\begin{abstract}
With this paper, the design of a biomimetic robotic squid (dubbed URSULA) developed for dexterous underwater manipulation is presented. The robot serves as a test bed for several novel underwater technologies such as soft manipulators, propeller-less propulsion, model mediated tele-operation with video and haptic feedback, sonar-based underwater mapping, localization, and navigation, and high bandwidth visible light communications. Following the finalization of the detailed design, a prototype is manufactured and is currently undergoing pool tests. 

\end{abstract}


\section{Introduction}
Fueled by the rapid population increase and the associated over-consumption of the 20th century, many experts believe that the world is heading for a resource crisis \cite{Wil08}. Higher demand on natural resources is forcing policy makers to consider increasing resource utilization, reducing waste, and improving recycling efficiency. In addition, the sustainable exploitation of the vast but mostly untouched oceans and seas is also another option. The marine environment is becoming a critical source of food, proteins, hydrocarbons and fossil fuels, alternative energy sources, mines and minerals. However, the underwater environment is not a suitable for humans to live or engage in any prolonged activity, making the utilization of unmanned underwater robots very attractive and often necessary for sub-sea exploration and intervention activities.  

The fleet of underwater robots used worldwide for commercial, scientific and military applications has grown significantly since the early 2000s. Despite their growing numbers, underwater unmanned robotic systems have changed very little in terms of their overall design principles and functional capabilities since they were first introduced in the 1970s. With the expansion of the application areas and the diversification of the tasks expected to be performed, current underwater robot designs will not be sufficient to meet the operational demands, in particular, in terms of underwater intervention \cite{Ble10}. To overcome this deficiency, there is a strong and urgent requirement to develop underwater robots with advanced manipulation and intervention capabilities \cite{Rib12}-\cite{Man18}.

\section{Requirements and Conceptual Design}

The primary objective of this work is to develop a robotic platform for dexterous underwater manipulation that combines the benefits of worker (remotely operated vehicles, ROV) and rover (autonomous underwater vehicles, AUV) robots. The proposed robot is to work close to the seabed, around both natural and artificial underwater structures. As a secondary objective, this robotic platform is designed to serve as a test bed for evaluating novel technologies that can be utilized in autonomous underwater systems.

\begin{figure}[b]
\centerline{\includegraphics[width=85mm]{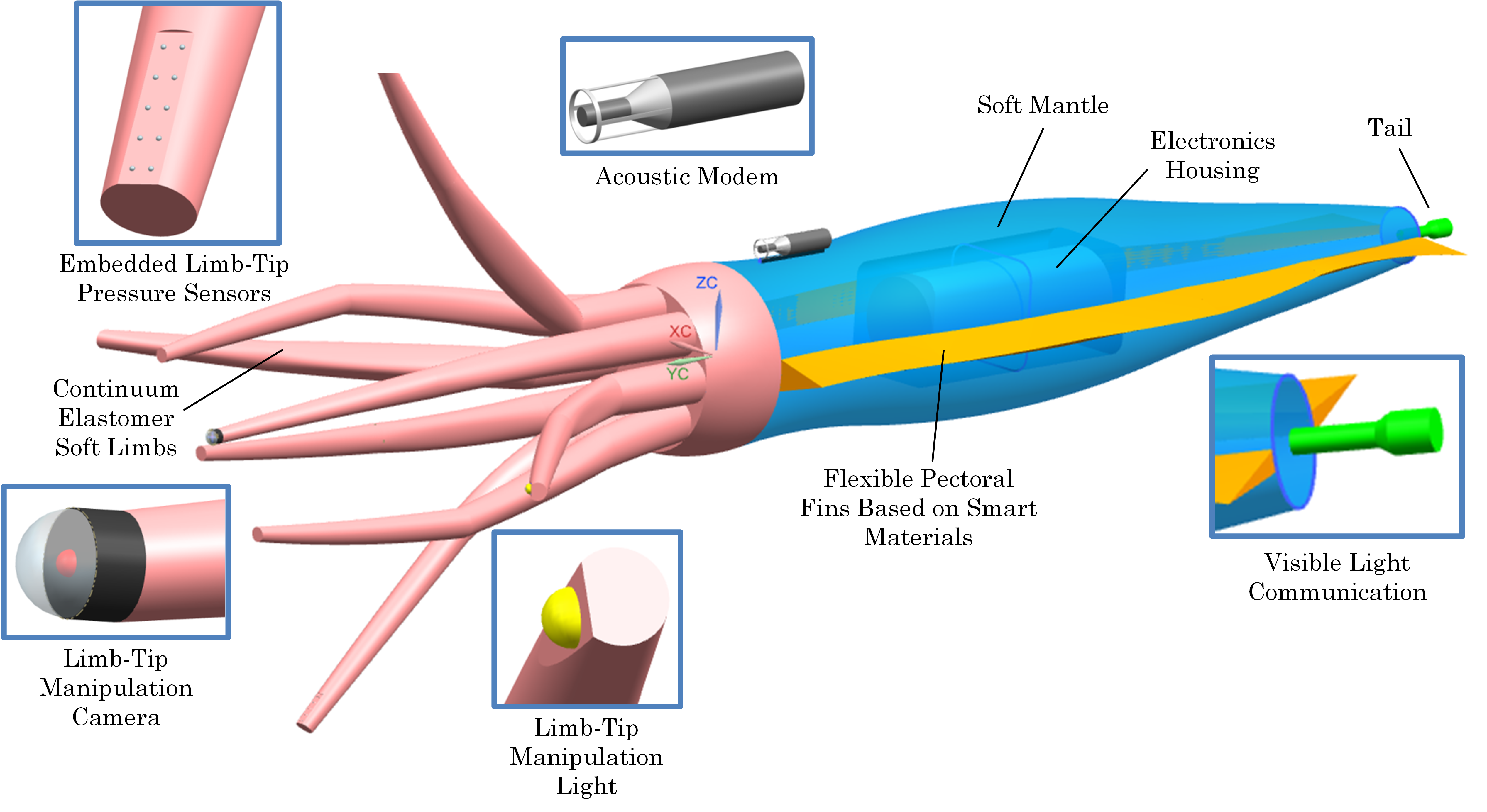}}
\caption{The conceptual design of URSULA.}
\label{fig: ConceptMechDesign}
\end{figure}

To this end, a cuttlefish or squid inspired biomimetic robot, very dissimilar to existing underwater robots is envisioned. The robot is designed to be compact, light-weight, and easily transportable to and deployable from offshore or coastal platforms without requiring specialized equipment or excessive manpower. Due to the risk of entanglement with marine growth and other underwater obstacles, high manoeuvrability and omni-directionality should be achieved with propeller-less propulsion systems. The robot must employ a sufficient number of soft limbs for performing tasks such as reaching, grasping, and retrieval. These limbs should be tele-operated and feature haptic feedback. A high-bandwidth wireless communication system is to be incorporated to provide video feedback to the operator during tele-operation. The initial maximum operating depth of the robot is not expected to exceed 100 m. The resulting robot concept, dubbed URSULA ({\textbf U}nmanned {\textbf R}obotic {\textbf S}quid for {\textbf U}nderwater and {\textbf L}ittoral {\textbf A}pplications) is depicted in Fig. \ref{fig: ConceptMechDesign}.   

\section{Detailed Design}

The finalized design of URSULA is presented in this section under three subsections that encompasses the mechanical, electronic, and software design.

\subsection{Body and Mechanical Design}\label{Sec: Mech}

\begin{figure}[b]
\centerline{\includegraphics[width=85mm]{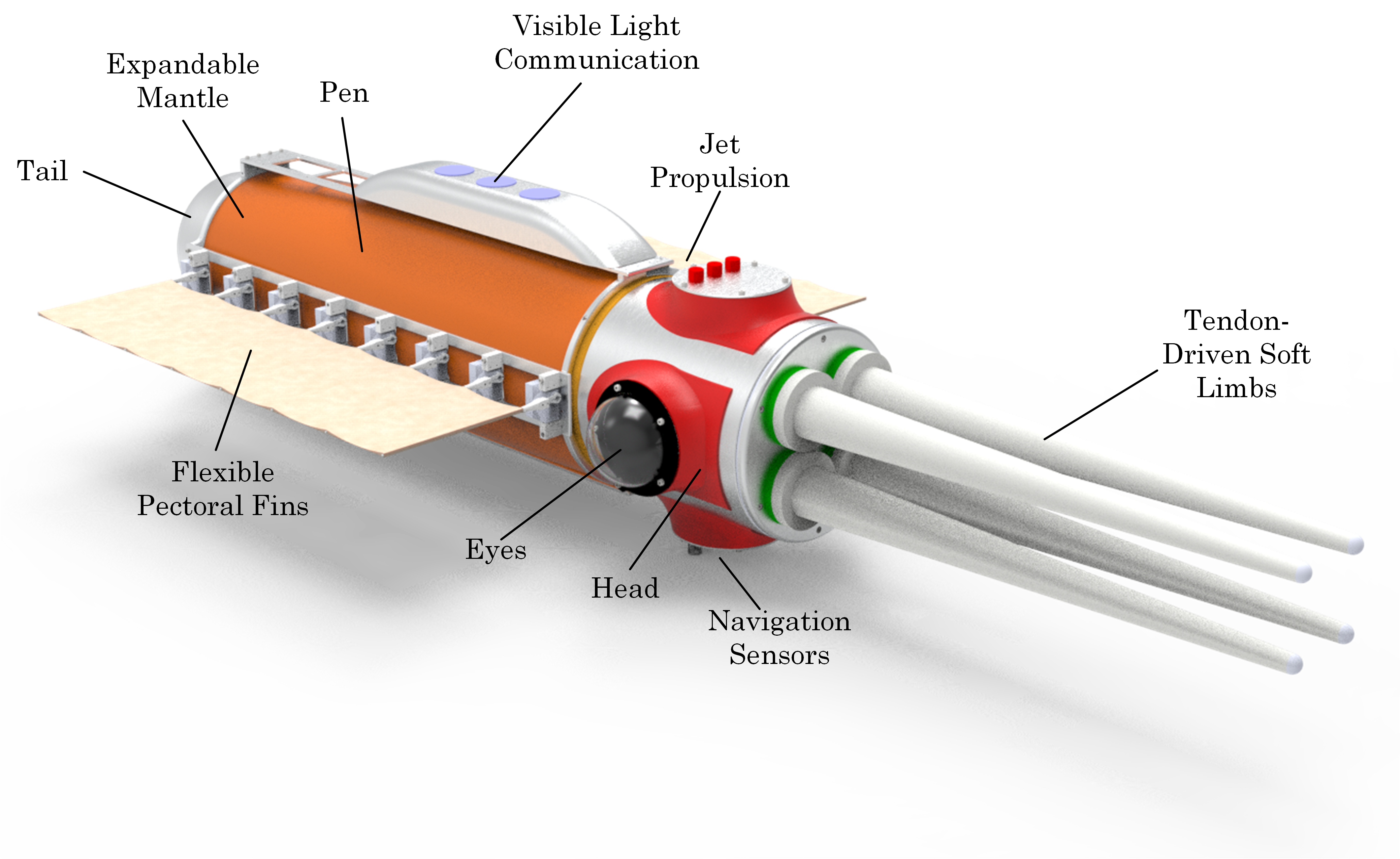}}
\caption{The solid model of the finalized design of URSULA with labeled main components.}
\label{fig: DetailMechDesign}
\end{figure}

Inspired by the common squid and cuttlefish, URSULA is comprised of 7 main body parts: 1) soft limbs, 2) head, 3) pen, 4) expandable mantle, 5) tail, 6) visible light communication system, and 7) flexible pectoral fins (see Fig. \ref{fig: DetailMechDesign}). The overall robot length (excluding the limbs) is less than 1200 mm with a maximum diameter of 250 mm. The limbs are 600 mm long, bringing the total mass of the robot to around 30 kg. The head, pen, and tail form the rigid, watertight components of the main body. Four tendon driven soft limbs are attached to the head. The head houses the tendon actuators as well as several navigation sensors. The pen bridges the head to the tail, functioning as an internal shell that hosts the electronics housing and batteries. The connector for the optional umbilical cable and other waterproof connections are located at the tail. Computational fluid dynamics based numerical analyses was conducted to achieve a hydrodynamically satisfactory design with optimal component placements.

The most critical subsystem of URSULA is the manipulation system that is comprised of four soft, tendon actuated robotic limbs protruding from the head and the associated actuation, sensing, and control elements. Two of these limbs are designated as arms and are designed to provide functionality such as reaching, gripping, and pulling to perform manipulation tasks in the form of probing, grasping, and retrieval. These arms incorporate embedded tactile sensors at their tips utilized for haptic feedback during teleoperation. The remaining two limbs are designated as tentacles and host an embedded underwater camera and lighting system at their distal ends. The motion of the camera tentacle can be mirrored on to the light tentacle to effectively illuminate the workspace of the arms. The tentacles are fully functional limbs and can also be used for other manipulation tasks such as anchoring the robot. 

\begin{figure}[b]
\centerline{\includegraphics[width=85mm]{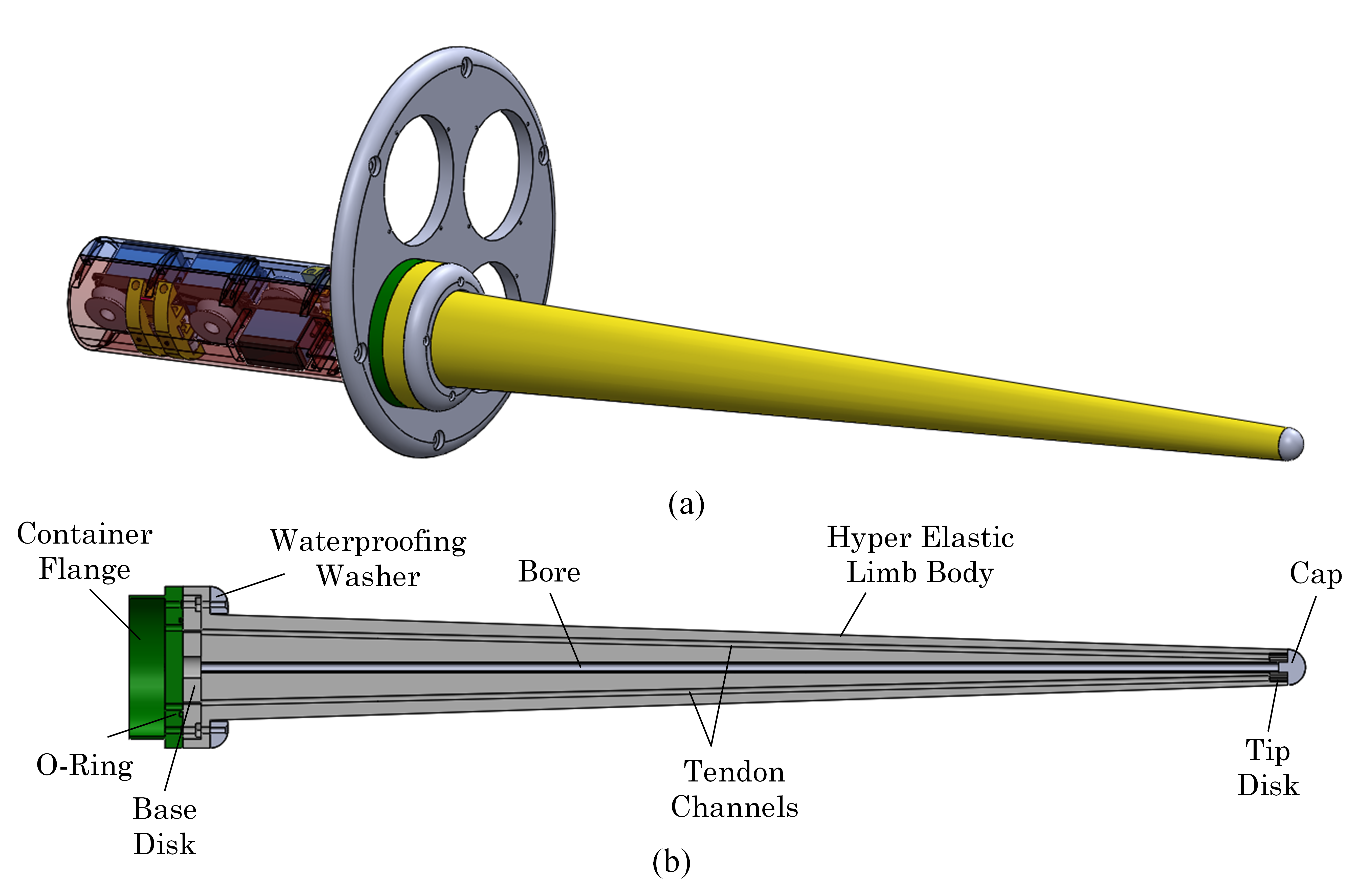}}
\caption{URSULA limbs; (a) The 3-D solid model of the limb system (showing only a single limb for clarity) and (b) the
cross-sectional cut of the limb labeled with the main parts.}
\label{fig: LimbModel}
\end{figure}

The limbs are designed as thin and slender frustums with a length of 600 mm as shown in Fig. \ref{fig: LimbModel} (a). Due to the size limitations of URSULA, the diameter at the base of the limbs (where the limbs are attached to the head) is 60 mm and reduces linearly to 20 mm at the tip. The limbs are actuated by a set of tendons channeled longitudinally through the body of the limbs (see Fig. \ref{fig: LimbModel} (b)). Although 3-tendon configurations are investigated with promising initial results, a 4-tendon configuration, first presented in \cite{Ren12}, is preferred for the preliminary design due to the simpler shape control strategies associated with the decoupling of the tendon actuation. Thin channels made from soft silicone tubes are used to guide the tendons along the length of the limbs and provide a protective sheath to prevent the tearing of the soft limbs. The tendons are anchored to a rigid disk embedded into the limb at the tip and are connected to electro-mechanical actuators at the other end. Each container at the base of the soft limb compactly houses the 4 actuators and tendon tension measurement systems \cite{Gur24}. Another rigid disk embedded at the base of the limb serves as the mechanical interface to 80 mm diameter flanges used to connect the limbs to the bow plate of the head. Aside from the tendon channels, an additional larger silicone tube is placed centrally along the spine of the arm and serves as the cable channel for the camera and lighting system. 

The bow and stern, port and starboard pectoral fins form the main actuation system of the robot. The design allows for optional port and starboard central fins to be added to the robot, if necessary. These central fins not only increase thrust, but can also be exploited to generate more complex undulations. The fins are attached to the two lateral rails on the port and starboard sides of the robot that run along the length of the pen, outside of the mantle. The fins are actuated either in flapping or standing wave (SW) mode, or in undulating or traveling wave (TW) mode to generate the necessary thrust. Another mode where the fins act as control surfaces is also envisioned but not yet studied in detail. The fins are designed to swivel 180$^{\circ}$. This wide motion range, coupled with the placement and actuation modes of the fins result in an omni-directional vehicle. The jet propulsion system is designated as the secondary propulsion system to complement the fins and is located towards the front part of the pen, inside of the expandable mantle. Sea water is pumped into the double-walled soft mantle. This causes the mantle to expand inwards increasing water pressure between the mantle and pen. When sufficient water pressure is obtained a solenoid valve is opened forcing the water out through a steerable nozzle.

The lateral rails are complemented with a pair of identical dorsal and ventral rails which are used to attach several navigational sensors, the visible light communication system, and potential payloads. The visible light communication system, in the form of a dorsal fin, provides a high-bandwidth, bilateral wireless communication interface required for the haptic teleoperation of the limbs. 

\subsection{Electronic Subsystem Design}
The onboard electronic subsystems include the mission computer, power distribution system, limb, fin and jet propulsion systems, visible light and other wireless communication systems, navigation system, perception system, as well as various sensor systems. A custom main board designed around the Raspberry Pi CM4 serves as the mission computer to handle command, control and communications. The main board communicates through various serial ports (e.g., RS485, RS422, RS232), an 10/100 Mbit ethernet port and a single USB2.0 port suitable for multiplexing. Also included on the main board is a ethernet switch for connecting various peripheral devices. 

URSULA can be powered either from onboard batteries (a pack of Li-Ion 6S4P batteries providing 25.2 V with 14 Ah capacity) or from a 400 V, 3 kW surface source via an umbilical cable. An isolated DC/DC convertor on the custom power distribution board reduces the incoming 400 V feed to the standard 28 V output. Separate DC/DC convertors are incorporated into the power distribution system to further reduce the voltage to levels required by various subsystems. Power monitoring capabilities are added at different levels for endurance estimation as well as subsystem and component level fault detection. The surface power source doubles as a charger to the onboard battery system.

A basic sensor package that consists of an IMU, GNSS and depth sensor, and a DVL is utilized for navigation. The GNSS sensor is used for absolute localization when the robot is at the surface. The design can also accommodate an optional perception system, that is comprised of a 3-D stereo vision camera, a sonar, and a separate perception computer for image processing. The Jetson Nvidia Nano GPU board is chosen as the perception computer for its ability to effectively execute GPU accelerated image processing and machine learning algorithms. The stereo vision system is included to primarily assist with manipulation and related tasks while the sonar is utilized for navigation. 

Both the limb and fin actuators are controlled using a custom designed driver board designed for the STM32F303 microcontroller. Each electronic card can control up to 8 actuators. Integrated analog-to-digital convertors are used to digitize sensor measurements (e.g., tendon tension or motor current) necessary for precise control of the actuators. The limbs and fins require two cards each. The jet propulsion system pump is driven by a single, separate custom designed electronic card with the STM32F303 microcontroller. 

During limb teleoperation, communication can be performed over ethernet through the umbilical cable or through the visible light communications system. LTE and WiFi wireless communication is also available when the robot is at the surface. A list of selected off-the-shelf electronic components used on URSULA is presented in Table \ref{tab: elecComp}.   

\renewcommand{\arraystretch}{1.5} 

\begin{table}[htbp]
\centering
\caption{Selected off-the-shelf electronic components and devices used on URSULA.}
\begin{tabular}{|c|c|c|c|}
\hline
\textbf{No.} & \textbf{Subsystem} & \textbf{Component} & \textbf{Hardware} \\
\hline
1 & Main Board & Mission Computer & Raspberry Pi CM4 \\
\hline
2 & \multirow{4}{*}{Navigation} & IMU & Movella / XSens-Mti7 \\
\cline{1-1} \cline{3-4}
3 & & GNSS & U-Blox ZED-F9P RTK \\
\cline{1-1} \cline{3-4}
4 & & Depth Sensor & Blue Robotics Bar30 \\
\cline{1-1} \cline{3-4}
5 & & DVL & WaterLinked A50  \\
\hline
6 & \multirow{3}{*}{Perception} & Computer & Nvidia Jetson Nano \\
\cline{1-1} \cline{3-4}
7 & & 3-D Vision Sensor & StereoLabs ZED \\
\cline{1-1} \cline{3-4}
8 & & Sonar & Blue Robotics Ping360 \\
\hline
9 & \multirow{2}{*}{Communication} & WiFi & Infineon CYW4345545 \\
\cline{1-1} \cline{3-4}
10 & & LTE & Quectel EC25 Mini PCIe \\
\hline
11 & \multirow{2}{*}{Power} & Power Supply & MeanWell CSP-3000-400 \\
\cline{1-1} \cline{3-4}
12 & & DC/DC Conv. & Cincon CFB750-300S28 \\
\hline
13 & \multirow{3}{*}{User Interface} & \multirow{3}{*}{Haptic Devices} & Novint Falcon \\
\cline{1-1} \cline{4-4}
14 & & & Force Dimension omega.6 \\
\cline{1-1} \cline{4-4}
15 & & & Geomagic Touch(X) \\
\hline
\end{tabular}
\label{tab: elecComp}
\end{table}

\subsection{Software Architecture}
The software of URSULA consists of the dry-end and wet-end modules. On the wet-end side, a ROS1 based mission control software handles tasks such as navigation, control of the fin, jet, and limb actuators, power monitoring and management, communications, and if active, stereo and sonar perception. 

The software is structured around several designated functional and operational modes. The robot can operate in either an explorer (EXP) or an intervention (INT) mode. In the EXP mode, URSULA operates similar to conventional AUV's, generally (but not necessarily) navigating autonomously (AUTNAV mode) in long endurance and passive missions such as exploration, search, mapping and monitoring. The robot switches to the INT mode when performing its primary task of underwater manipulation. In this INT mode, the robot can be completely manually tele-operated (MANCON mode), effectively becoming a conventional ROV. However, for reducing the workload of the operator during intervention, a semi-autonomous posture control (SAUTPOS) mode based on whole body control strategies \cite{Kha04} is also being developed. Finally, in terms of power supply and communications, either a tethered (TET) or a wireless (NOWIRE) mode can be selected. 

Mechanics-based dynamic models of continuum, hyper-elastic limbs are generally very complex and developing accurate and efficient analytic inverse models for such soft robots remain illusive. Thus, numerical or approximate schemes are frequently exploited for this purpose \cite{Eme24}.  As an alternative, data-driven machine learning approaches are utilized in this research for the tip position and motion control of the limbs using tendon displacements or tension forces. More specifically, a transformer-based architecture \cite{Alk23a} and reinforcement learning methods \cite{Alk23b} are exploited for way-point and time-dependent trajectory type motion, respectively. The current work on control of limbs using machine learning techniques is expected to expand to shape and contact force control of the soft limbs. 

A model-mediated, haptic feedback teleoperation scheme is being developed for manually controlling the limbs in the INT mode (see and \cite{Eme22} and \cite{Cez23} for preliminary simulation results). The primary approach to involves utilizing limb embedded RGB cameras and tactile sensors for haptic feedback enhanced bilateral teleoperation. To improve teleoperation performance and robustness, an alternative approach based on the 3-D stereo vision system is also being developed. In this approach, depth and shape data obtained from the stereo camera at short range is utilized to extract develop an abstract and virtual representation of the 3-D environment around URSULA. This model is used to estimate contact forces at the dry-end. Discrepancies between the actual environment and the abstract model are eliminated based on actual tactile measurements made on the limbs.     

Two basic approaches for guidance and navigation are implemented. First, the nonlinear, adaptive, dynamic line-of-sight controller presented in \cite{Agu07} is utilized for autonomous way-point navigation. A more recent guidance and control method \cite{Kog16} based on the Udwadia-Kalaba approach to constrained dynamic systems is for preferred for following time-dependent trajectories. Both approaches are implemented in a simpler 2-D setup that decouples surge, sway, and yaw motion from the heave motion, and completely ignores roll and pitch dynamics. In addition to these well known guidance and navigation methods, a sonar-based underwater simultaneous localization and mapping (SLAM) approach is proposed and evaluated through simulations. This method exploits maps of artificial underwater landmarks and utilizes machine learning-based object detection and tracking to estimate the current underwater location. Similar to terrestrial SLAM, when fused with dead-reckoning and other available position estimates, the proposed underwater SLAM approach appears to be a viable alternative to acoustic localization methods in use today \cite{Ese23}.      

\begin{figure}[t]
\centerline{\includegraphics[width=85mm]{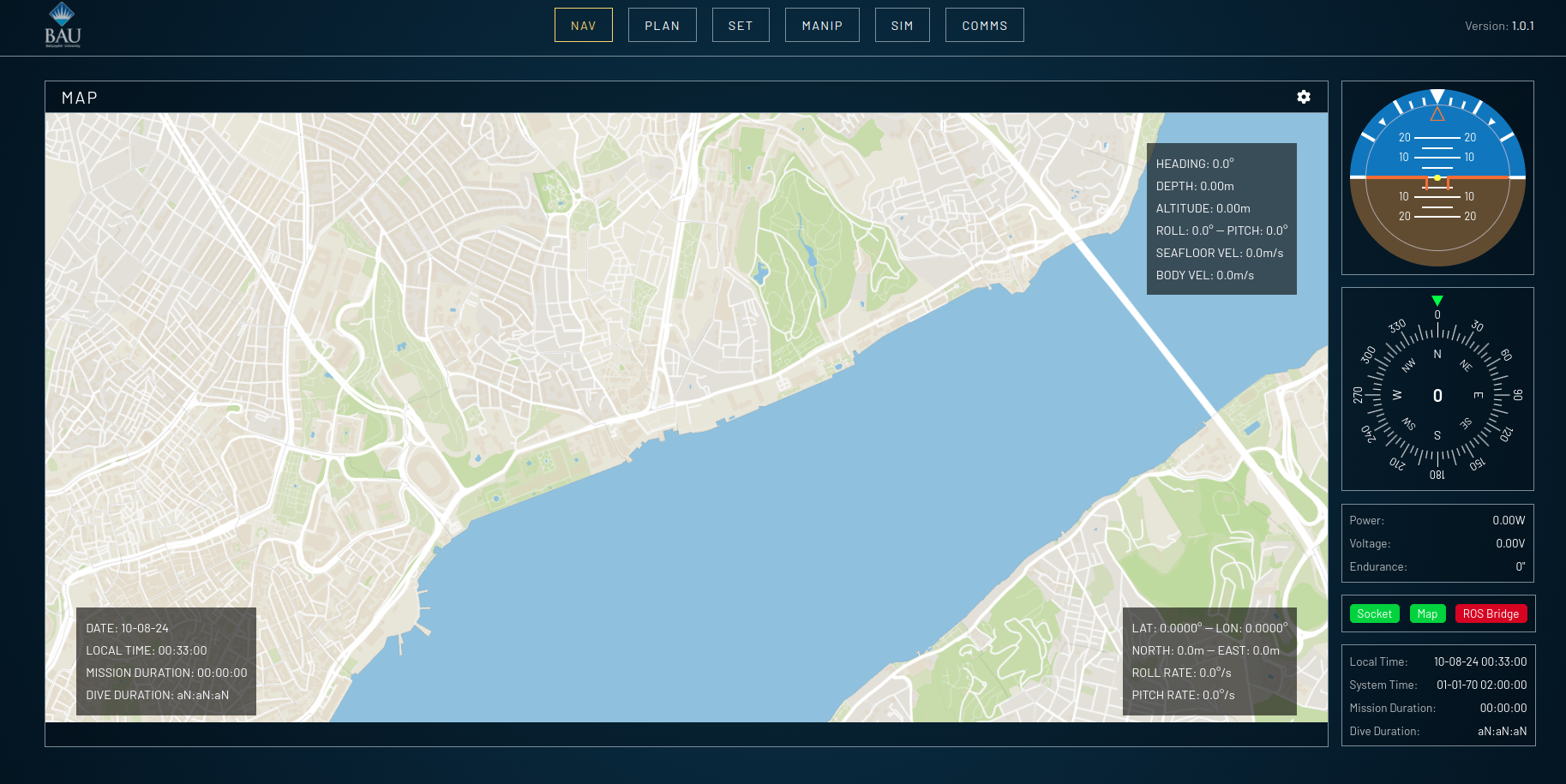}}
\caption{An early version of the navigation module of the web-based UI developed for URSULA.}
\label{fig: WebUI}
\end{figure}

The primary top-side software is a user interface (UI) with navigation, mission planning, situational awareness and manipulation modules (see Fig. \ref{fig: WebUI}). A web-based UI is preferred for accessibility, extensibility, and scalability. In the future, it is envisioned that this UI is to be expanded in terms of capabilities to control and command heterogeneous fleets and swarms of unmanned underwater (UUV), surface (USV), and aerial vehicles (UAV) as well as other fixed and mobile unmanned assets. It should be noted that tasks with high computational burden such as limb motion planning and control are also performed at the top-side.

\begin{figure}[b]
\centerline{\includegraphics[width=85mm]{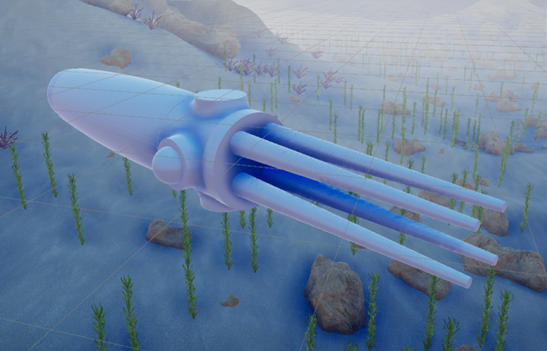}}
\caption{A view of URSULA in the simulation environment developed for evaluating XR enabled dexterous underwater manipulation \cite{Muf24}.}
\label{fig: UnitySim}
\end{figure}

A separate underwater simulator based on the Unity gaming engine (see Fig. \ref{fig: UnitySim}) is developed to conduct preliminary studies on the effectiveness of haptic feedback and extended reality (XR) in performing difficult underwater manipulation tasks (some preliminary results are discussed in \cite{Muf24}). This simulator is designed to seamlessly integrate with the URSULA user interface and replace the actual robot.      

\section{Prototype Development and Tests}\label{sec: prototype}
After the completion of the detailed design, URSULA subsystems were manufactured and tested in a laboratory setting. The main water-tight body components were constructed from Al5754 and Al6082 Aluminum alloys to obtain a light-weight and corrosion resistant structure.  

\begin{figure}[htbp]
\centerline{\includegraphics[width=85mm]{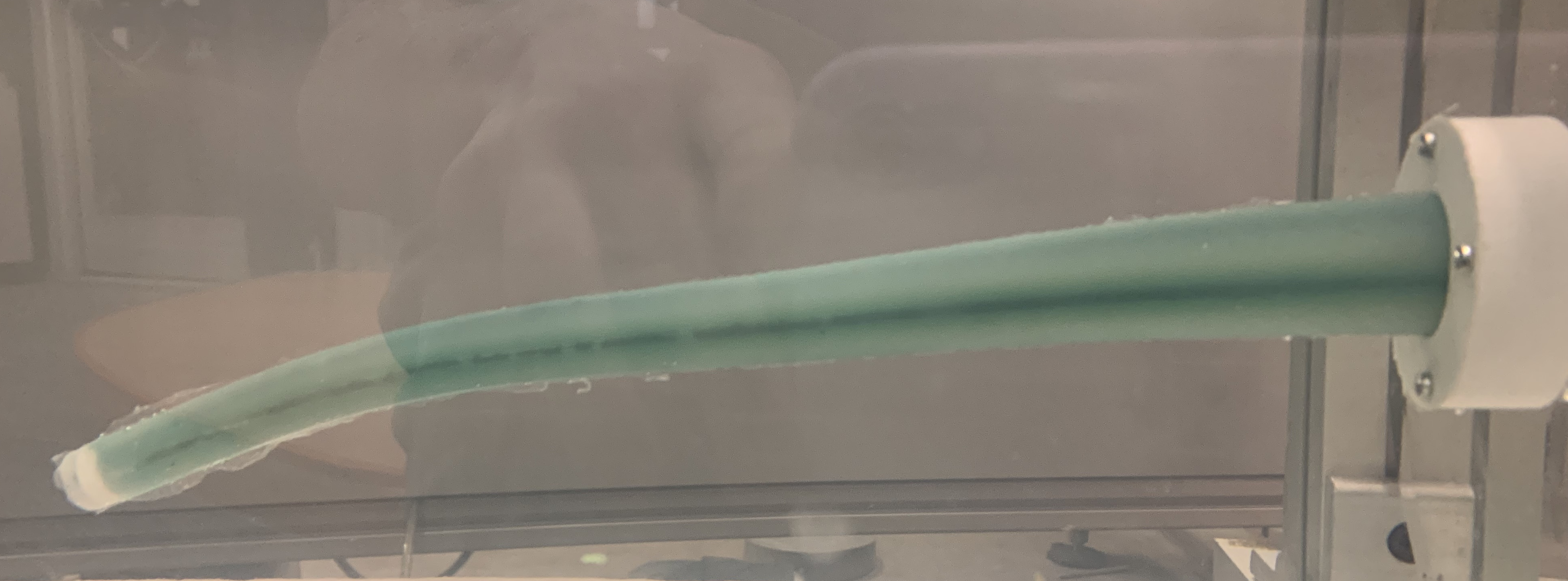}}
\caption{A half-scale soft limb floating in a test tank.}
\label{fig: Limb}
\end{figure}

The hyper-redundant, soft robotic limbs are molded from high quality silicon rubber with very few 3-D printed rigid parts acting as anchor points. The use of soft and light-weight silicon rubber material not only enables the limbs to safely interact with the fragile underwater environment and handle delicate payloads, but also results in naturally buoyant limbs (see Fig. \ref{fig: Limb}). Laboratory tests conducted with the soft limb actuation and tendon tension measurement systems, the fin and the jet pump system, as well as the visible light system are previously reported in \cite{Gur24}.  

\begin{figure}[t]
\centerline{\includegraphics[width=85mm]{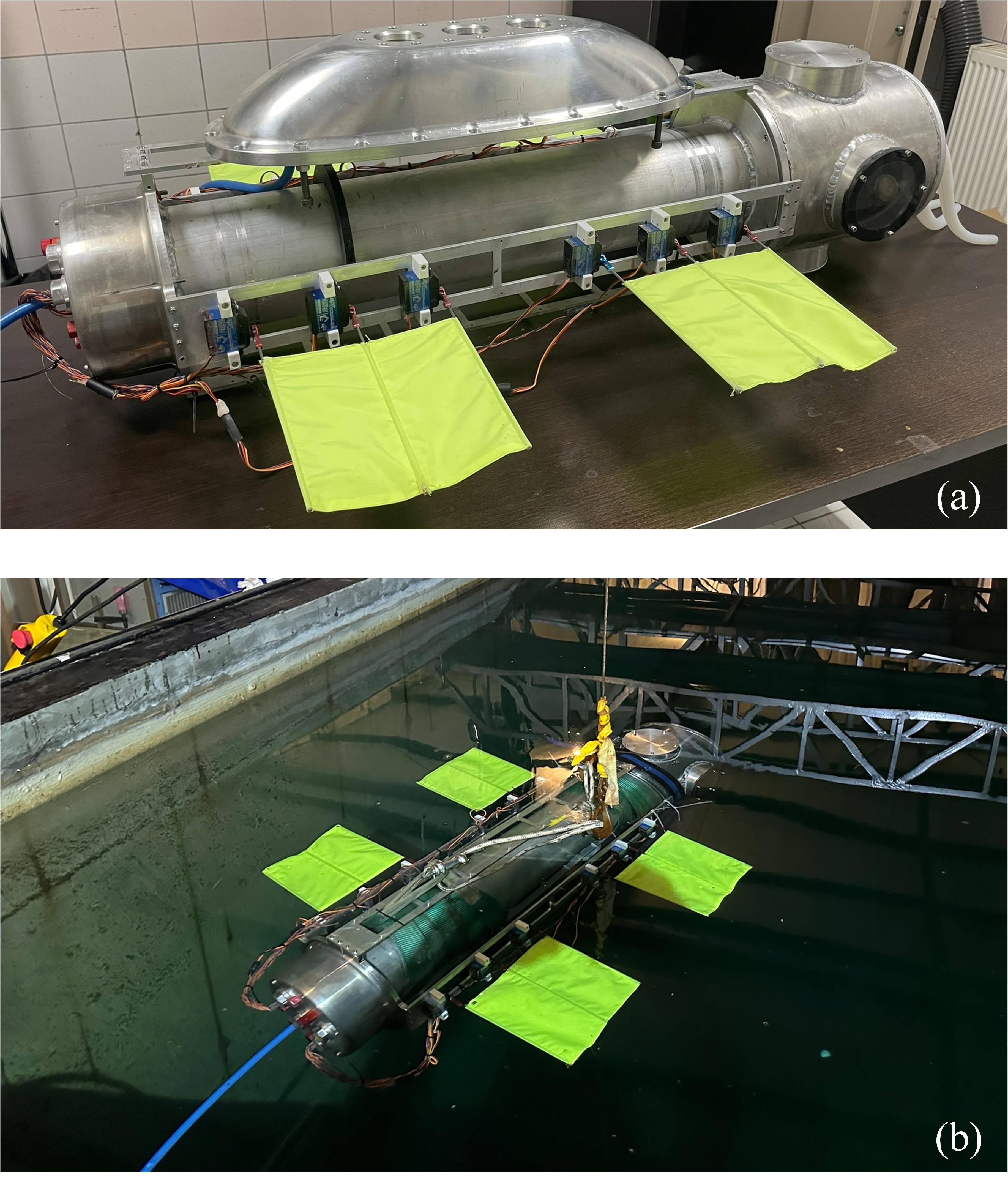}}
\caption{The first prototype of URSULA (a) fully assembled except for the mantle and (b) undergoing pool tests.}
\label{fig: Prototype}
\end{figure}

Upon completion of subsystem level laboratory test and validation efforts, the prototype robot body was manufactured and integrated with the electronics, manipulation, fin, and communication subsystems (see Fig. \ref{fig: Prototype} (a)). A safety switch and other critical safety-related measures were incorporated to the robot prior to the beginning of the first phase pool tests (see Fig. \ref{fig: Prototype} (b)). URSULA currently is deployed with half-scale limbs since the full-length limbs are still in the development phase. 

\section{Conclusions and Future Work}\label{sec: conc}

In this work, the design details of project URSULA, a squid-inspired biomimetic robot developed for dexterous underwater manipulation is presented. The proposed general purpose class robot has a total length and mass of 1800 mm and 30 kg, respectively. URSULA hosts several novel and emerging technological features such as soft robotic manipulators, propeller-less propulsion, model mediated tele-operation with video and haptic feedback, and high bandwidth visible light communication. Currently, the first prototype of the robot is undergoing pool tests as sub-systems are gradually being integrated and tested. In the meantime, work on a second prototype has begun. This second prototype will include an improved and lighter body design capable of reaching 100 m of depth, as well as modifications to sub-systems such as hydraulically actuated limbs and suckers for improved dexterity, fins without rigid components, a more compact visible light system and a ROS2 based software suite for commanding a heterogeneous swarm of networked aerial, surface, and sub-sea drones.

\end{document}